\documentclass[conference,10pt]{IEEEtran}
\IEEEoverridecommandlockouts
\usepackage{hyperref}
\usepackage{url}
\usepackage{tabularx}
\usepackage{amsmath,amssymb,amsfonts}
\usepackage[ruled,vlined,linesnumbered]{algorithm2e}
\usepackage{graphicx}
\usepackage{textcomp}
\usepackage{xcolor}
\usepackage{wrapfig}
\usepackage{etoolbox}
\usepackage{adjustbox}
\makeatletter
\patchcmd{\@makecaption}
  {\scshape}
  {}
  {}
  {}
\makeatother
\def\BibTeX{{\rm B\kern-.05em{\sc i\kern-.025em b}\kern-.08em
    T\kern-.1667em\lower.7ex\hbox{E}\kern-.125emX}}
\begin{document}

\title{AxFormer: Accuracy-driven Approximation of Transformers for Faster, Smaller and more Accurate NLP Models\\
}
\author{\IEEEauthorblockN{Amrit Nagarajan, Sanchari Sen\textsuperscript{\textsection}, Jacob R. Stevens\textsuperscript{+} and Anand Raghunathan}
\IEEEauthorblockA{\textit{School of ECE, Purdue University} \\
\{nagaraj9, sen9, steven69, raghunathan\} @ purdue.edu}}

\maketitle

\renewcommand\thefootnote{\textsection}
\footnotetext{Currently at IBM T.J. Watson Research Center (sanchari.sen@ibm.com).}

\renewcommand\thefootnote{+}
\footnotetext{Currently at Meta Platforms Inc. (jrstevens@fb.com).}

\setlength{\belowcaptionskip}{-1.75pt}
\setlength{\abovecaptionskip}{-1.75pt}

\begin{abstract}
Transformers have greatly advanced the state-of-the-art in Natural Language Processing (NLP) in recent years, but present very large computation and storage requirements. We observe that the design process of Transformers (pre-train a foundation model on a large dataset in a self-supervised manner, and subsequently fine-tune it for different downstream tasks) leads to task-specific models that are highly over-parameterized, adversely impacting both accuracy and inference efficiency. We propose {\it AxFormer}, a systematic framework that applies accuracy-driven approximations to create optimized transformer models for a given downstream task. AxFormer combines two key optimizations --- accuracy-driven pruning and selective hard attention.
Accuracy-driven pruning identifies and removes parts of the fine-tuned transformer that hinder performance on the given downstream task. Sparse hard-attention optimizes attention blocks in selected layers by eliminating irrelevant word aggregations, thereby helping the model focus only on the relevant parts of the input. In effect, AxFormer leads to models that are {\em more accurate, while also being faster and smaller}. Our experiments on GLUE and SQUAD tasks show that AxFormer models are up to 4.5\% more accurate, while also being up to 2.5$\times$ faster and up to 3.2$\times$ smaller than conventional fine-tuned models. In addition, we demonstrate that AxFormer can be combined with previous efforts such as distillation or quantization to achieve further efficiency gains. Code is available at \url{https://github.com/amrnag/Specialized-Transformers}.

\end{abstract}

\section{Introduction}
Transformer models~\cite{DBLP:conf/naacl/DevlinCLT19,radford2019language,brown2020language} have revolutionized the field of Natural Language Processing (NLP), greatly advancing the state-of-the-art in many NLP tasks. Models that achieve good performance on these tasks are of high practical significance, finding their place in commercial applications such as social media monitoring (sentiment analysis), AI chat assistants (question answering), automated summarization tools (analyzing sentence similarity), {\em etc.} Therefore, there is a strong interest in creating more accurate and efficient models for these tasks.

Transformers are first pre-trained on very large datasets, enabling them to capture rich linguistic knowledge and gain a deep understanding of the structure of the target language. Subsequent fine-tuning refines this knowledge for a specific downstream task by training a task-specific final layer. However, we find that the current method of pre-training and fine-tuning transformers has two major drawbacks. First, large pre-trained transformers are highly over-parameterized for downstream tasks, especially since many of these tasks have very limited training data. This leads to unstable models~\cite{DBLP:journals/corr/abs-2002-06305} with sub-optimal generalization ability~\cite{DBLP:conf/nips/MichelLN19}. Second, these large fine-tuned models present high computation and storage requirements for inference. This problem is compounded by the trend towards larger models, driven by the need for higher accuracy and tackling more complex tasks. For instance, increasing the number of parameters from 1.5B to 175B enabled a reduction in perplexity for Language Modeling (on the Penn Treebank dataset) from 35.8 in GPT-2 (2019) to 20.5 in GPT-3 (2020). In this work, we address both these challenges, taking advantage of the over-parameterized nature of pre-trained models to create individually optimized models for the different downstream tasks that are smaller, faster and more accurate.

We propose AxFormer, a framework that combines two key optimizations -- accuracy-driven pruning and selective replacement of ``soft" self-attention with ``hard" self-attention. Accuracy-driven pruning identifies and removes elements (we define elements as parameters grouped at different levels of granularity,  {\em i.e.}, self-attention blocks, feed-forward neural network blocks, attention heads and neurons) of the transformer that hinder performance on a specific downstream task, with the goal of maximizing accuracy. In contrast to prior pruning methods that prune elements with little-or-no impact on the network output, the proposed method prunes elements that actually have a considerable impact on the output, leading to the highest positive impact on accuracy. In order to reduce the large pruning space, we consider the different elements of the fine-tuned transformer for pruning in a hierarchical manner, starting with entire self-attention or feed-forward neural network blocks, followed by attention heads, and finally individual neurons.

The core of the transformer is self-attention, where each token in the input builds its representation based on the extent of attention it places on all the other tokens. However, we observe that in some layers, restricting the attention span of each token to only focus on the relevant tokens leads to better information flow inside the model. AxFormer identifies the appropriate layers and replaces the ``soft" self-attention with ``hard" self-attention in these layers. AxFormer does not require any additional re-training or fine-tuning (thereby enabling it to scale to large transformer models), and can be applied in a plug-and-play manner to any Transformer model that is fine-tuned for any downstream task.

We summarize our main contributions as follows: 
\begin{itemize}
\item We introduce {\it AxFormer}, a framework that optimizes transformer models through approximate computing for specific downstream tasks. 
\item We propose accuracy-driven pruning to identify and eliminate elements that are harmful to performance on the downstream task.
\item We propose the selective replacement of soft self-attention with hard attention in certain layers, helping the model to focus only on the relevant parts of the input to build better representations.
\item Across a suite of different transformer networks and downstream tasks, we demonstrate that AxFormer models are consistently more accurate and stable, while also being significantly faster and smaller than their (conventional) fine-tuned counterparts.
\end{itemize}

\section{Related Work}
We discuss relevant previous effors by categorizing them along three directions - task-agnostic optimizations, task-specific optimizations and pruning.

\noindent\textbf{Task-agnostic optimizations.} Given the popularity of transformer models, several techniques have been proposed to overcome their computational and memory challenges, for efficient inference. A vast majority of these works introduce task-agnostic optimizations by pre-training efficient models from scratch. Some notable examples include DistilBERT \cite{DBLP:journals/corr/abs-1910-01108} and MobileBERT \cite{DBLP:conf/acl/SunYSLYZ20}, which use knowledge distillation to train smaller and faster networks using the original network as a teacher. LayerDrop \cite{DBLP:conf/iclr/FanGJ20} randomly drops layers during pre-training, thereby enabling their dropping during inference. Lite Transformer \cite{DBLP:conf/iclr/WuLLLH20} uses Long-Short Range Attention to speed up the self-attention operation. AlBERT \cite{DBLP:conf/iclr/LanCGGSS20} uses factorized embeddings and cross-layer parameter sharing. Q8BERT \cite{DBLP:journals/corr/abs-1910-06188} quantizes all weights and activations in the model to 8-bit integers. Using DistilBERT and Q8BERT as examples, we demonstrate that our techniques are complementary to these works, and can be applied to derive benefits over and beyond them. 

\smallskip
\noindent\textbf{Task-specific optimizations.} QBERT  \cite{DBLP:conf/aaai/ShenDYMYGMK20} and TinyBERT \cite{DBLP:journals/corr/abs-1909-10351} perform task-specific quantization and knowledge distillation, but the gain in efficiency comes at the cost of degradation in accuracy on the downstream task. In contrast, AxFormer models are simultaneously and more efficient.

\smallskip
\noindent\textbf{Pruning techniques.} Pruning has been applied to various classes of neural networks, including transformers (\cite{DBLP:conf/iclr/FrankleC19,DBLP:conf/iclr/MolchanovTKAK17,sajjad2020poor}), and pruning strategies have been explored to identify weights and parameters that the output is least sensitive to (\cite{DBLP:conf/nips/HouHSJCL20,DBLP:conf/icml/YeGNZKL20}). In contrast, AxFormer identifies and prunes parameters that have the most detrimental effect on the output, leading to simultaneous gains in accuracy and efficiency. Using the popular Lottery Ticket Hypothesis \cite{DBLP:conf/iclr/FrankleC19} as an illustrative example, we demonstrate that our accuracy-driven pruning method is complementary to previously proposed pruning techniques, and hence, they can be combined to produce even more efficient models for a given accuracy constraint.

\section{The AxFormer Framework}

The AxFormer framework for producing optimized transformer models for specific downstream tasks is illustrated in Algorithm \ref{alg:alg5}. It performs two key optimizations: (1) Identify and prune elements that hinder performance on the given downstream task (lines 16-25), and (2) Selectively replace soft self-attention with hard self-attention to help the model focus only on the relevant parts of the input (lines 9-15). We describe these steps in the following subsections.

\begin{algorithm}[ht]
\small
\caption{The AxFormer Framework}
\label{alg:alg5}
  \DontPrintSemicolon
  \SetKwInOut{Input}{Input}
 \SetKwInOut{Output}{Output}
  \Input{ Fine-tuned (for the given downstream task) Transformer T, Validation set D}
  \Output{ Optimized AxFormer model for the given downstream task T}
  
  \SetKwFunction{FMain}{$analyze\_element$}
  \SetKwProg{Fn}{Function}{:}{}
  \Fn{\FMain{element E}}{
     $T_{pruned} = $ T $ - $ E \;
   	    $New\_Loss$ = $Evaluate(T_{pruned},D)$ \;
   	    \If{$New\_Loss < Min\_Loss$ and $num\_samples\_helped > 0.5\times cardinality(D)$}
   	      {
   	       $Min\_Loss$ = $New\_Loss$ \;
   	       T = $T_{pruned}$
   	       }
  }

    $Min\_Loss$ = Evaluate (T,D) \\
    Q = $Order\_elements\_for\_inspection(T,D)$ \\
    \For{each layer L in T}
   	   {Replace soft self-attention in L with hard self-attention \\
   	    $New\_Loss$ = $Evaluate(T,D)$ \\
   	    \If{$New\_Loss < Min\_Loss$ and $num\_samples\_helped > 0.5\times cardinality(D)$}
   	      {
   	       $Min\_Loss$ = $New\_Loss$ \\
   	       }
   	      \Else
   	      { 
   	       Restore soft self-attention in L \\
   	      }
   	   }
    
    \While {Q is not empty}
    {
    TrialElement = Q.pop() \\
    $analyze\_element$(TrialElement)\\
   	\If{TrialElement has not been pruned from T and $New\_Loss < 1.1 \times Min\_loss$}
   	{
   	 \If{TrialElement is an attention block}
   	   {
   	   \For{each attention head h in TrialElement}
   	   {
   	    $analyze\_element$(h)\\
   	   }
   	   }
   	   \ElseIf{TrialElement is a feed-forward block}
   	   {
   	   \For{each neuron w in TrialElement}
   	   {
   	    $analyze\_element$(w)\\
   	   }
   	   }
   	}
   }
    return T
\end{algorithm}

\subsection{Accuracy-driven pruning}
The problem of identifying an optimal set of elements to prune in order to maximize accuracy is challenging, and this is especially true for transformers for two reasons. First, transformers have billions of parameters, leading to an enormous pruning search space. Second, the iterative \{approximate, fine-tune, approximate\} cycle used in previously proposed pruning methods (to recover accuracy loss from pruning) does not work for large transformers during fine-tuning, since they very quickly overfit the limited training data for the downstream tasks (usually within 2-3 epochs). We address both these challenges through the use of a hierarchical greedy algorithm that does not require any additional training or fine-tuning (lines 16-25 in Alg. \ref{alg:alg5}). To determine the significance of each transformer element for the given downstream task, we first fine-tune the original transformer model for the given downstream task to obtain the baseline loss. Then, for the element under consideration in each iteration of the framework, we compute the loss of the current transformer model with the element removed. We prune the element under consideration if the validation loss when it is removed is less than the minimum loss seen thus far during the optimization process, since the goal is to find a model with minimum loss. As a result, contrary to previously proposed pruning methods, accuracy-driven pruning can be seen as a form of training, since the objective of accuracy-driven pruning -- minimizing loss on the (validation) dataset -- is exactly the same as the objective of training (see Appendix C). Also, in order to prevent overfitting to the validation set, we introduce a generalization constraint in addition to the aforementioned loss condition. This constraint ensures that an element is pruned only if it decreases the loss of more than half the total number of samples in the validation set (computed by $num\_samples\_helped$ function in Alg. \ref{alg:alg5}). Therefore, elements are pruned only if a majority of the samples in the validation set benefit from their removal, resulting in improved generalization performance. If the loss with the element removed is slightly greater than the minimum loss seen so far, we inspect the element at a finer granularity, and prune only parts of the element that hinder performance (rather than pruning the entire element). 

\subsubsection{\textbf{Hierarchical processing of elements}} 
It is computationally prohibitive to analyze every single parameter in large transformers using the method described in Alg. \ref{alg:alg5}. Since the framework iterates through the entries of the queue sequentially, its efficacy is dependent on both the total number of elements under consideration, and the time required to analyze each element. We take advantage of the inherently hierarchical structure of transformers and consider the elements in a hierarchical manner, ordered by increasing granularity. Specifically, we analyze entire feed-forward and self-attention blocks first, and inspect them at finer granularity (attention heads and neurons) only when required. Through this ordering, we are able to quickly eliminate large numbers of parameters from further consideration. In addition, due to the over-parameterized nature of transformers, it is likely that time-consuming blocks are pruned from the transformer earlier in the process, thereby speeding up future iterations of the framework. For example, eliminating a single feed-forward block in the BERT-Base model removes 5.6\% of all parameters under consideration, and speeds up future iterations by 1.15$\times$. To further reduce the number of elements under consideration, we also dynamically remove elements if they are encompassed by a high-importance block. For example, if a given self-attention block is determined to be of high importance (the validation loss with the block removed is much greater than the minimum loss seen so far), we remove all heads within that block from further consideration.

\subsubsection{\textbf{Creating an ordered queue of elements for inspection}} Since our framework performs greedy pruning of highly over-parameterized models, the order in which elements are inspected has a significant impact on the quality (accuracy, speed and size) of the final AxFormer model. Some elements may appear to be harmful for the downstream task (especially in early iterations of the framework), but this often ends up being an artifact of over-parameterization. As a result, when elements are not carefully ordered for inspection, our framework lands in bad local minima of the validation loss function, where loss cannot be reduced further through pruning. Our solution to this problem utilizes the unique linguistic properties captured by the different transformer layers \cite{DBLP:conf/acl/JawaharSS19} to guide the ordering of elements for inspection (line 8 in Alg. \ref{alg:alg5}) For example, it was found that BERT captures phrase-level information in the lower layers, mapping related tokens together. The lower layers also capture surface features, while the middle layers capture syntactic features and higher layers capture semantic features. It was also observed that BERT requires deeper layers only when long-range dependency information is required. Different tasks require different types of linguistic knowledge. For example, sentiment analysis requires only local context, and long-range information often ends up confusing the model, since sentiments often change rapidly; it is also unlikely that syntactic and semantic information are needed (see Appendix A). Hence, we place the final layer at the front of the queue, and work our way backwards towards the first layer, since blocks in the final layers are more likely to hinder performance on sentiment analysis. This ordering ensures that elements that are pruned early in our framework (when the model is most over-parameterized) do not lead the system into bad local minima, since elements that are pruned are expected to hinder performance based on prior insights into the working of transformers. We note that the use of prior knowledge about different tasks is only a heuristic to minimize the overheads of the AxFormer framework. All the exact results in this paper can be obtained without assuming any knowledge about the downstream task (but at the cost of additional overheads), using the procedure described in Appendix B.

\subsection{Selective use of hard self-attention}
\begin{figure}[htb]
\centerline{\includegraphics[width=\linewidth]{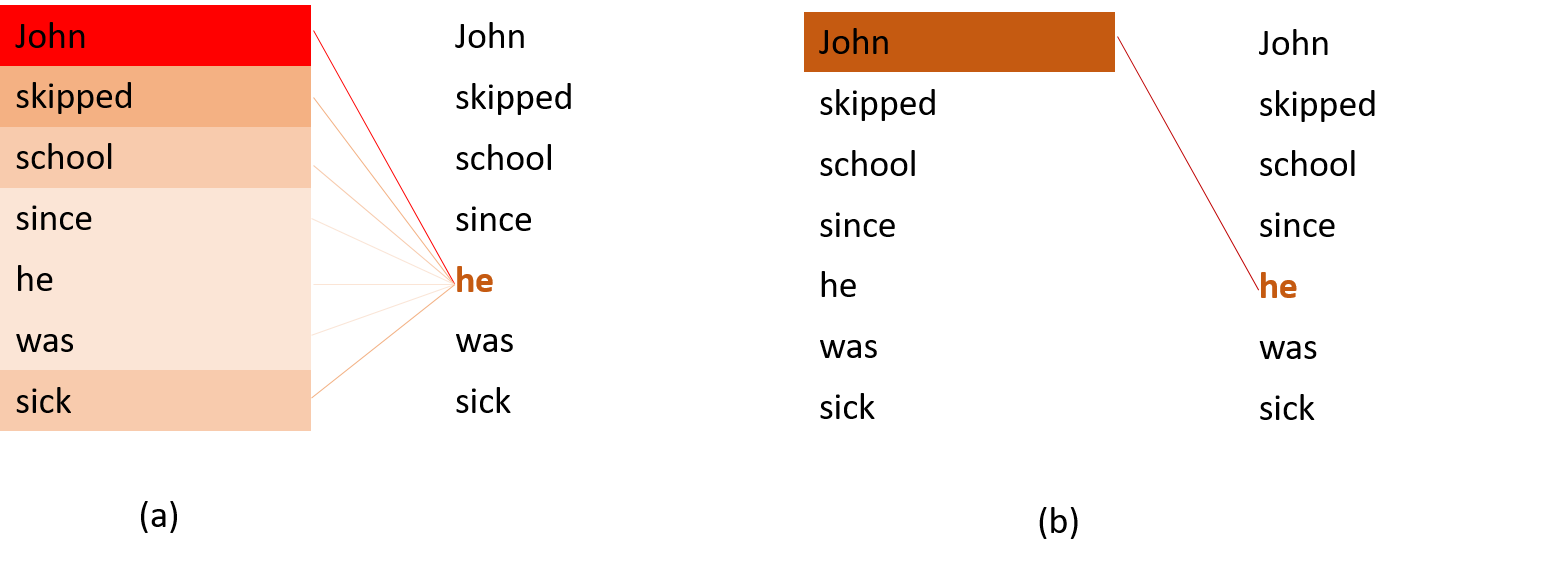}}
\caption{\textbf{Illustrations of (a) Traditional ``soft" self-attention and (b) ``Hard" self-attention for the word ``he".} In hard self-attention, ''he" concentrates all of its focus on ``John", while in soft self-attention, ``he" places a small amount of attention on the other irrelevant words also. Therefore, hard self-attention helps build a better representation for ``he". }
\label{fig:hattn}
\end{figure}

 In traditional transformer architectures, the self-attention operation computes the attention scores of each token in the input sequence with all the other tokens. These attention scores, after passing through the softmax operation, are used to build the new representation for the token based on the extent of attention it places on the other tokens. The use of ``soft" attention enables end-to-end training of the transformer. However, we observe that considering only the top-K attention scores for each token (K=30 is optimal for all of our studied tasks) in selected layers after training and fine-tuning is completed helps the transformer focus only on the relevant parts of the input sequence (Fig.~\ref{fig:hattn}). This leads to better information flow inside the model, thereby improving accuracy on downstream tasks (especially those involving sequence classification and question answering). This also introduces a large amount of activation sparsity that sparse  accelerators can exploit for faster inference, and helps alleviate the critical memory bottleneck in transformers. We note that replacing soft attention with hard attention in all layers (especially the deeper layers) leads to loss of accuracy, necessitating its selective use. In order to identify the layers that benefit from hard attention, we replace the soft attention with hard attention in each layer (one by one), and use it only when it leads to smaller validation loss. 

\section{Experimental Results}

We implemented our techniques within Huggingface’s Transformers library in PyTorch \cite{DBLP:journals/corr/abs-1910-03771}. We used Intel AI's NLP Architect for experiments on Q8BERT. The experiments were performed on a GeForce RTX 2080 Ti GPU with 11GB memory. All results are reported on the test set, averaged across 10 runs with random seeds after 3 epochs of fine-tuning, unless otherwise specified. We randomly sample $15\%$ of the training set with class balance, and use it as the validation set in the AxFormer framework. Thus, the test set is not used in AxFormer. 

\subsection{AxFormer models are faster, smaller and more accurate than conventional fine-tuned models}  
We present results on GLUE \cite{DBLP:conf/iclr/WangSMHLB19}, a set of Language Understanding tasks, and SQUADv1.1 \cite{DBLP:conf/emnlp/RajpurkarZLL16}, a Question Answering task, in Table \ref{tab:table30}. For GLUE, we present results on the test set using the GLUE evaluation server to obtain the scores \cite{DBLP:journals/corr/abs-2002-02925}, and for SQUAD, we present results on the dev set. AxFormer models are {\em up to 4.5\% more accurate, while also being up to 2.5$\times$ faster and up to 3.2$\times$ smaller than the baseline}. In addition, AxFormer models show substantial improvements over Q8BERT-base (which is already 4$\times$ smaller than BERT-base due to the use of 8-bit integer quantization) and DistilBERT-base (which is already 60\% faster and smaller than BERT-base). {\em AxFormer versions of Q8BERT-base and DistilBERT-base models exceed the accuracy of BERT-base models, while being up to 3.7$\times$ faster and 12.1$\times$ smaller than BERT-base}. Therefore, AxFormer is complementary to, and can be used in conjunction with, prior quantization and distillation methods. The benefits of AxFormer are further analyzed in Appendix C.

\begin{table*}[htbp]
\centering
\fontsize{5}{6}\selectfont
\caption{\textbf{Accuracy, speedup and compression of AxFormer models on GLUE and SQUAD v1.1 tasks.} We report Matthews correlation for CoLA, Pearson Correlation for STS-B and accuracy for all other tasks. We report only ``matched'' accuracy for MNLI and the Exact Match score for SQUAD.  Speedup and Compression are reported over the non-AxFormer baselines for DistilBERT and Q8BERT, and not over BERT-base.}
\resizebox{\textwidth}{!}{%
\begin{tabular}{@{}cccccccccccc@{}}
\hline
{} & \textbf{SQUAD} & \textbf{CoLA}& \textbf{MNLI}& \textbf{MRPC}& \textbf{QNLI}& \textbf{QQP} & \textbf{RTE}& \textbf{SST-2}& \textbf{STS-B}& \textbf{WNLI}& \textbf{Average} \\  \hline   
BERT-Base & 81.97 & 51.38 & 85.37 & 83.11 & 90.74 & 89.76 & 62.65 & 95.94 & 83.8 & 64.1 & 78.88  \\ 
\textbf{AxFormer} & \textbf{83.88} & \textbf{54.43} & \textbf{86.63} & \textbf{85.88} & \textbf{91.17}& \textbf{91.12}& \textbf{65.81}& \textbf{96.7}& \textbf{85.08}& \textbf{65.1}& \textbf{80.58} \\ 
Speedup & 1.38X & 1.98X & 1.21X & 1.61X & 1.16X & 1.49X & 2.2X & 1.32X & 1.71X & 2.51X & 1.67X \\
Compression & 1.52X & 2.88X & 1.6X & 2.02X & 1.31X & 1.48X & 2.69X & 1.44X & 1.38X & 3.18X & 1.95X \\
\hline
Q8BERT-Base & 81.55&	50.5&	84.73&	82.05&	90.19&	89.65&	61.26&	94.32&	83.09 & 61.21 & 77.86 \\
\textbf{AxFormer} & \textbf{82.52} &	\textbf{54.12} &	\textbf{85.95} &	\textbf{84.77} &	\textbf{91.08} &	\textbf{91.44} &	\textbf{65.8} &	\textbf{96.02} &	\textbf{84.14} &	\textbf{63.1} &	\textbf{79.73} \\
Speedup & 1.31X & 	2.28X & 	1.08X & 	1.57X& 	1.3X & 	1.26X & 	2.28X & 	1.15X & 	1.58X &  	2.44X & 	1.63X \\
Compression &   1.41X &  3.02X	&   1.24X	&  1.47X	&  1.2X	&  1.38X	&  2.74X	&  1.21X&  1.46X & 2.82X&   1.8X\\
\hline
DistilBERT-Base &	79.94&	50.5&	83.04&	81.07&	89.72&	88.92&	62.01&	93.88&	83.49&	61.3&	77.39\\
\textbf{AxFormer} &  \textbf{81.91} &	\textbf{54.08} &	\textbf{85.47} &	\textbf{85.02} &	\textbf{91.01} &	\textbf{89.2} &	\textbf{64.5} &	\textbf{95.48} &	\textbf{84.16} &	\textbf{63.3} &	\textbf{79.41} \\
Speedup &	1.21X & 	1.84X &	1.09X &	1.34X &	1.07X &	1.21X &	1.94X &	1.08X &	1.31X & 	2.31X &	1.44X \\
Compression	&1.28X 	&1.96X	& 1.14X	&1.41X	& 1.08X	& 1.14X	&1.7X	& 1.17X	&1.28X 	& 1.98X	&     1.41X\\
\hline
XLNet-Base&	82.28&	51.98&	85.93&	83.68&	91.01&	89.83&	62.01&	95.88&	84.12&	64.8&	79.15\\
\textbf{AxFormer} & \textbf{83.93} &	\textbf{54.28} &	\textbf{86.8} &	\textbf{85.92} &	\textbf{91.78} &	\textbf{91.12} &	\textbf{65.78} &	\textbf{96.98} &	\textbf{85.46} &	\textbf{65.1} &	\textbf{80.71} \\
Speedup&	1.48X& 	2.06X &	1.38X &	1.98X &	1.47X &	1.78X &	2.11X &	1.59X &	1.93X & 	2.46X &	1.83X \\
Compression&	 1.55X 	&2.84X	& 1.72X	& 2.12X	& 1.56X	& 1.6X	&2.48X	&1.71X	& 1.65X & 2.98X	&      2.02X\\
\hline

\end{tabular}
}%
\label{tab:table30}
\end{table*}

\subsection{AxFormer models are less sensitive to random seed initialization} 
Previous research \cite{DBLP:journals/corr/abs-2002-06305} has shown that the random seed used to determine the initialization of the task-specific layer and the order of training data for the downstream task has a significant impact on the quality of the fine-tuned models. When the amount of training data is small (which is the case with most downstream NLP tasks), this effect becomes more pronounced, evidenced by the fact that WNLI, which has the least amount of training samples (634 training samples), exhibits highest variance across runs. Therefore, in order to find a model with high accuracy on a downstream task, multiple models need to be created (using different random seeds), and evaluated. In Table \ref{tab:table31}, we demonstrate that AxFormer models exhibit significantly less sensitivity to random seeds than their non-AxFormer counterparts, thereby greatly increasing the odds of finding  ``good'' models in fewer iterations.  

\begin{table}[htbp]
\caption{\textbf{Sensitivity to random seeds.} Results reported are averaged across the GLUE tasks and SQUAD on the Base models of each transformer.}

\begin{tabularx}{\linewidth}{@{} *5{>{\centering\arraybackslash}X}@{}}

\hline
\textbf{Transformer} & \textbf{Average variance across tasks}& \textbf{Maximum variance (single task)}& \textbf{Average of maximum task scores}& \textbf{Average of minimum task scores}\\ 
\hline

BERT & 1.48 & 4.86 & 80.05 & 76.53  \\

\textbf{AxFormer} & \textbf{0.86} & \textbf{1.95} & \textbf{82.02} & \textbf{79.93}  \\
\hline

Q8BERT & 1.61 & 4.62 & 79.01 & 75.29  \\

\textbf{AxFormer} & \textbf{1.1} & \textbf{2.02} & \textbf{81.14} & \textbf{79.58}  \\
\hline

DistilBERT & 1.29 & 3.85 & 78.78 & 74.89  \\

\textbf{AxFormer} & \textbf{0.83} & \textbf{1.78} & \textbf{80.96} & \textbf{79.59}  \\ 
\hline

\end{tabularx}
\label{tab:table31}
\end{table}

\subsection{Large transformers benefit greatly from AxFormer}
Current state-of-the-art transformer networks (such as GPT-3 \cite{brown2020language}) have hundreds of billions of parameters. Model sizes are also expected to grow further in the future as increasing the number of parameters has been shown to improve performance. This makes it computationally challenging to train transformers as well as perform inference using them. Recent research \cite{DBLP:conf/icml/LiWSLKKG20} has shown that larger models converge in a significantly smaller number of training iterations than smaller models, and hence they train faster in spite of requiring more time per iteration. However, larger models are significantly slower than smaller models at inference time. In Table \ref{tab:table32}, we demonstrate that AxFormer applied to larger transformers achieves much higher accuracy while being comparable in speed to AxFormer models of smaller transformers. This demonstrates the scalability of AxFormer to larger models.


\begin{table}[htbp]

\caption{\textbf{AxFormer models of BERT-base and BERT-large.} Results reported are averaged across the GLUE tasks and SQUAD.}

\begin{tabularx}{\linewidth}{>{\centering\arraybackslash}X >{\centering\arraybackslash}X >{\centering\arraybackslash}X}

\hline
\textbf{Transformer} & \textbf{Accuracy}& \textbf{Samples/second}\\
\hline
BERT-Base & 78.88 & 0.21 \\
\textbf{AxFormer}  & \textbf{80.58} & \textbf{0.12}  \\
\hline
BERT-Large & 80.98 & 0.45 \\
\textbf{AxFormer}  & \textbf{82.89} & \textbf{0.15}  \\
\hline
\end{tabularx}
\label{tab:table32}
\end{table}

\subsection{Accuracy-driven pruning is complementary to conventional pruning techniques} 
\begin{wrapfigure}{r}{0.6\linewidth}

\includegraphics[width=\linewidth]{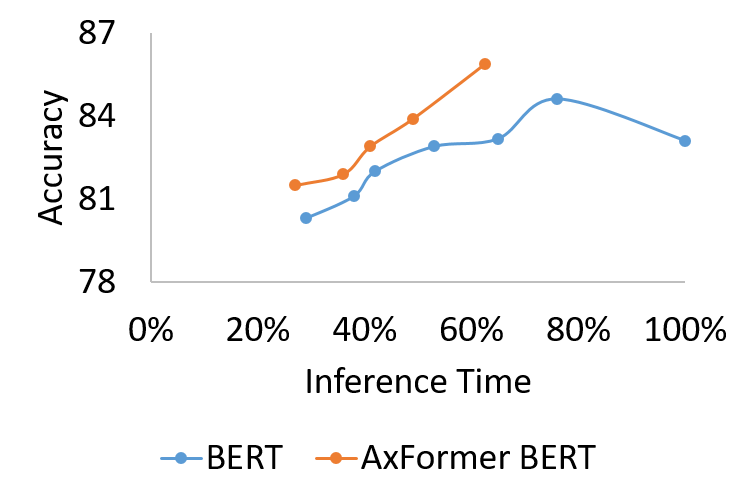}

\caption{\textbf{The winning tickets of conventional fine-tuned and AxFormer BERT-Base on MRPC.}}
\label{fig8}
\end{wrapfigure}
Accuracy-driven pruning identifies and prunes harmful elements of the transformer for the downstream task at hand. Techniques that prune elements that have minimal impact on the output are complementary to our techniques, and we demonstrate this using the popular Lottery Ticket Hypothesis \cite{DBLP:conf/iclr/FrankleC19}, which finds sparse sub-networks that can be trained to match the performance of the large network. The Lottery Ticket Hypothesis has been successfully applied to transformers (\cite{DBLP:conf/acl/LiangZCJLHZC20,DBLP:conf/emnlp/PrasannaRR20}). We demonstrate that the winning tickets of the AxFormer model are consistently more accurate than the winning tickets of conventional fine-tuned models for a comparable inference time (Figure \ref{fig8}).




\subsection{AxFormer provides insights into the working of transformers} 
\begin{figure*}[htbp]
\centerline{\includegraphics[width=0.9\linewidth]{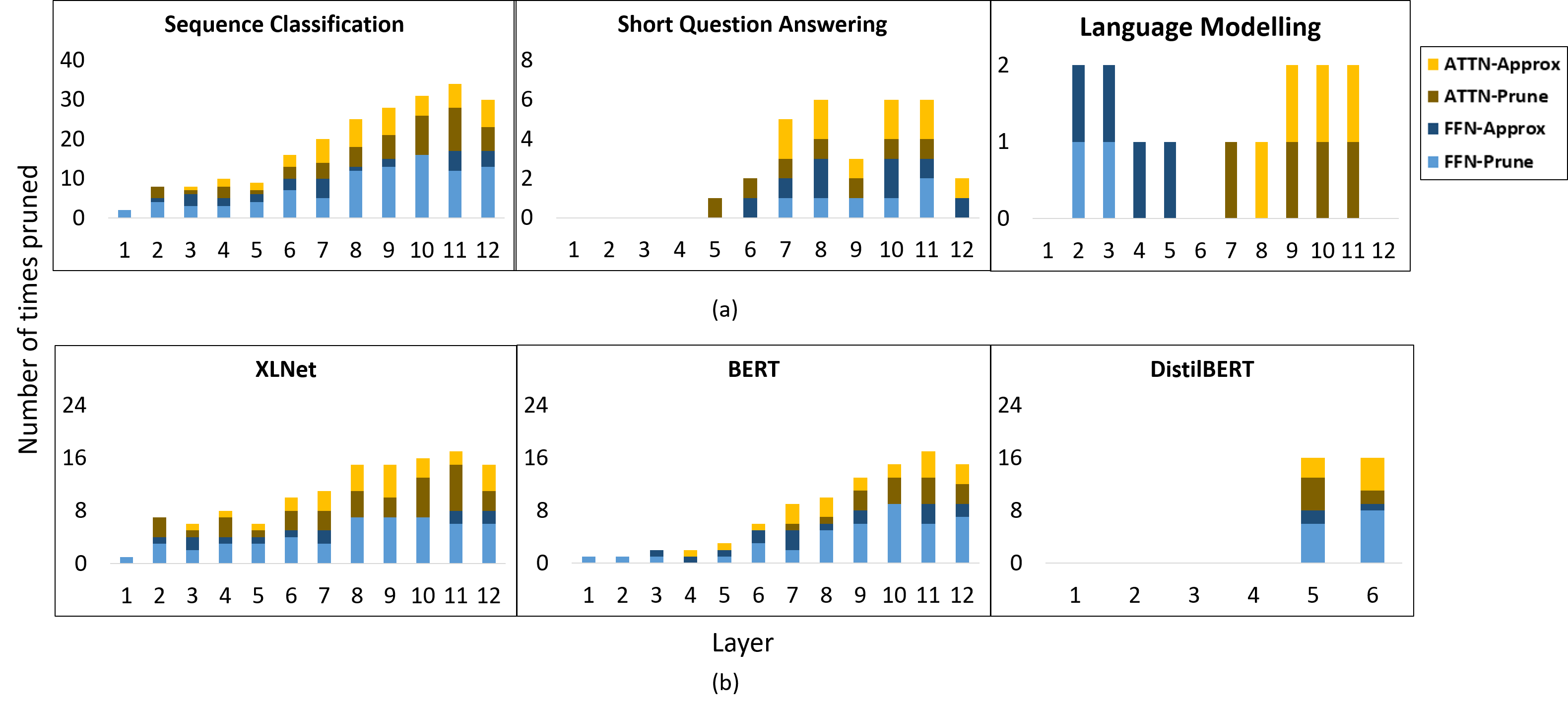}}
\caption{\textbf{Elements pruned for (a) downstream tasks and (b) Transformer architectures across GLUE and SQUAD in our most accurate models for each task.} Here, ATTN-Approx means only certain attention heads inside the attention block are pruned, and FFN-Approx means only certain neurons inside the feed-forward block are pruned. }
\label{fig9}
\end{figure*}

We analyze which elements of the transformer are pruned for different downstream tasks using different transformer models (Fig.~\ref{fig9}). We find that the differences in importance of elements are more pronounced across different tasks than across different models. For example, for sentiment analysis, long-range dependency information is not required, and often ends up confusing the model. Hence, for all models fine-tuned for sentiment analysis, we observe that components in later layers (closer to the output) are more likely to be pruned. This is not the case with Language Modeling tasks (predicting masked words in text), where longer attention spans are required. Across models, we only observe subtle differences. For example, we find that XLNet (auto-regressive) is able to learn important task-specific information earlier than BERT (non auto-regressive), similar to the observation made in \cite{sajjad2020poor}. Hence, we are able to drop more components (in earlier layers) in XLNet than in BERT, leading to more efficient models for inference. In DistilBERT (a distilled model), we find that there is a clear demarcation in linguistic knowledge across layers due to the reduced capacity of the model. This is evidenced by the fact that elements in the top four layers are never pruned across all Language Understanding tasks, while the boundaries are more soft in the original models. We also observe that Hard Attention is most useful in the lower layers, where phrase-level information is captured, for all the models and tasks.   

\subsection{Accuracy-driven pruning using previously proposed importance estimation techniques leads to less accurate models}
L1-norm and magnitude-based pruning techniques are designed to prune elements that have minimal impact on the output. Hence, they are not effective at pruning elements that have detrimental impact on output. Gradient-based methods such as Taylor expansion are not reliable for increasing accuracy on the downstream tasks, since transformers cannot be repeatedly fine-tuned to recover the accuracy losses from approximating the model (they very quickly overfit the limited training data for the downstream tasks). In addition, while knowledge distillation is complementary to our method (as evidenced by our results on DistilBERT), we show that the accuracy gains from AxFormer are significantly higher than those from regularized knowledge distillation proposed in \cite{DBLP:conf/nips/HouHSJCL20}. Super-tickets \cite{DBLP:conf/acl/LiangZCJLHZC20} demonstrate that accuracy can be improved by finding winning lottery tickets at low levels of pruning. In Table \ref{tab:table38}, we find that alleviating the over-parameterization issue by pruning elements that hinder performance (and the selective use of hard attention) leads to more accurate models than pruning elements that have least impact on output in the original model.

\begin{table}[htbp]
\caption{\textbf{Results of pruning with previously proposed methods (MRPC with BERT-Base).} For L1-norm, magnitude and Taylor (absolute gradient of loss), we prune 5\% of the least important weights at a time and record the test accuracy. We report the highest test accuracy seen in this process.}

\begin{tabularx}{\linewidth}{>{\centering\arraybackslash}X >{\centering\arraybackslash\hsize=.25\hsize}X}

\hline
\textbf{Pruning Method} & \textbf{Accuracy}\\
\hline
Baseline (No Pruning) & 83.11 \\

L1-Norm & 83.16  \\

Magnitude & 83.18 \\

Taylor Expansion (with additional fine-tuning) & 83.19 \\

Taylor Expansion (no additional fine-tuning) & 83.32 \\

Regularized Knowledge Distillation (DynaBERT) & 83.46 \\

Super-ticket & 84.62 \\

\textbf{Ours} & \textbf{85.88} \\
\hline
\end{tabularx}
\label{tab:table38}
\end{table}

\subsection{AxFormer runtime} 
The AxFormer framework does not require any additional training or fine-tuning iterations. It only requires multiple passes through a small validation set. Hierarchical processing ensures that the number of elements to inspect is greatly reduced. In addition, each pass is expected to become progressively faster, since the framework potentially prunes an element in each iteration, and our ordering of elements ensures that large, time-consuming blocks are pruned early in the process, making future iterations faster. Therefore, the runtimes of AxFormer are quite small compared to the time required for re-training or additional fine-tuning. We find that the average wall-clock time for AxFormer is only 7.6 minutes for BERT-base and 18.3 minutes for BERT-large on a single NVIDIA RTX 2080Ti GPU.

We also evaluate the heuristics used in our framework (Table  \ref{tab:table35}). When elements are not carefully ordered for inspection, we find that elements that are pruned in early iterations of the framework are pruned due to over-parameterization, and not because the linguistic knowledge contained in these elements is harmful/not useful for the task at hand. This leads the system into bad local minima, where the validation loss cannot be further reduced by pruning other elements. We observe that starting at the layer granularity leads to worse models than starting at the block granularity. This is because the attention and feed-forward blocks in each layer have vastly differently functionality, and hence, the effect of removing an attention block of relatively high significance is often masked by removing the feed-forward block that greatly hinders performance in the same layer (and vice-versa), causing the entire layer to be pruned. This also leads the system into bad local minima, thereby creating inferior models. To establish that our greedy approach combined with a global error bound does not lead to inferior models, we also experiment with an adaptive loss threshold. In particular, we use a very tight constraint when analyzing elements at coarse granularities, and relax the constraint as we move towards finer granularities. We again find that there is negligible change in the quality of the final model produced (the accuracy is slightly lower, possibly due to over-fitting to the validation set), but the AxFormer overheads are significantly larger. We hypothesize that a single global error bound is sufficient because we order the elements in such a way that for the given task at hand, we intuitively expect that the elements at the head of the queue are likely to be removed using the linguistic knowledge in different layers.

\begin{table}[htbp]
\caption{\textbf{[Right] Accuracy of the resulting models and runtime of AxFormer with different heuristics (MRPC with BERT-Base). } All heuristics use hierarchical processing of ordered elements with Selective Hard Attention, unless otherwise specified.}
\begin{tabularx}{\linewidth}{>{\centering\arraybackslash}X >{\centering\arraybackslash\hsize=.25\hsize}X >{\centering\arraybackslash\hsize=.25\hsize}X}
\hline
\textbf{Heuristic} & \textbf{AxFormer accuracy}& \textbf{Runtime (minutes)}\\
\hline
Randomly ordered elements & 83.42 & 7.2 \\

Start by inspecting layers & 83.44 & 5.8 \\

No Selective Hard Attention & 84.94 & 6.3 \\

Adaptive Threshold & 85.82 & 29.3 \\

\textbf{Ours} & \textbf{85.88} & \textbf{6.6}  \\
\hline
\end{tabularx}
\label{tab:table35}
\end{table}

\section{Conclusion}
We propose AxFormer, a framework to optimize  transformers for different downstream tasks. The framework identifies and prunes elements that hinder performance on the downstream task at hand. We also demonstrate the advantage of selectively using hard self-attention in selected layers to improve information flow. Using this framework, we produce models that are more accurate, while also being faster and smaller than conventional fine-tuned models. 

\smallskip
{\noindent \bf Acknowledgment:} This work was supported by C-BRIC, one of six centers in JUMP, a Semiconductor Research Corporation (SRC) program sponsored by DARPA.

\bibliographystyle{plain}
\bibliography{references.bib}

\begin{thebibliography}{10}

\bibitem{brown2020language}
Tom~B. Brown, Benjamin Mann, and Nick~Ryder et~al.
\newblock Language models are few-shot learners.
\newblock {\em CoRR}, 2020.

\bibitem{DBLP:conf/naacl/DevlinCLT19}
Jacob Devlin, Ming{-}Wei Chang, Kenton Lee, and Kristina Toutanova.
\newblock {BERT:} pre-training of deep bidirectional transformers for language
  understanding.
\newblock In {\em {NAACL-HLT} 2019}.

\bibitem{DBLP:journals/corr/abs-2002-06305}
Jesse Dodge, Gabriel Ilharco, Roy Schwartz, Ali Farhadi, Hannaneh Hajishirzi,
  and Noah~A. Smith.
\newblock Fine-tuning pretrained language models: Weight initializations, data
  orders, and early stopping.
\newblock {\em CoRR}, 2020.

\bibitem{DBLP:conf/iclr/FanGJ20}
Angela Fan, Edouard Grave, and Armand Joulin.
\newblock Reducing transformer depth on demand with structured dropout.
\newblock In {\em {ICLR} 2020}.

\bibitem{DBLP:conf/iclr/FrankleC19}
Jonathan Frankle and Michael Carbin.
\newblock The lottery ticket hypothesis: Finding sparse, trainable neural
  networks.
\newblock In {\em {ICLR} 2019}.

\bibitem{DBLP:conf/nips/HouHSJCL20}
Lu~Hou, Zhiqi Huang, Lifeng Shang, Xin Jiang, Xiao Chen, and Qun Liu.
\newblock Dynabert: Dynamic {BERT} with adaptive width and depth.
\newblock In {\em NeurIPS 2020}.

\bibitem{DBLP:conf/acl/JawaharSS19}
Ganesh Jawahar, Beno{\^{\i}}t Sagot, and Djam{\'{e}} Seddah.
\newblock What does {BERT} learn about the structure of language?
\newblock In Anna Korhonen, David~R. Traum, and Llu{\'{\i}}s M{\`{a}}rquez,
  editors, {\em {ACL} 2019}.

\bibitem{DBLP:journals/corr/abs-1909-10351}
Xiaoqi Jiao, Yichun Yin, Lifeng Shang, Xin Jiang, Xiao Chen, Linlin Li, Fang
  Wang, and Qun Liu.
\newblock Tinybert: Distilling {BERT} for natural language understanding.
\newblock {\em CoRR}, 2019.

\bibitem{DBLP:conf/iclr/LanCGGSS20}
Zhenzhong Lan, Mingda Chen, Sebastian Goodman, Kevin Gimpel, Piyush Sharma, and
  Radu Soricut.
\newblock {ALBERT:} {A} lite {BERT} for self-supervised learning of language
  representations.
\newblock In {\em {ICLR} 2020}.

\bibitem{DBLP:conf/icml/LiWSLKKG20}
Zhuohan Li, Eric Wallace, Sheng Shen, Kevin Lin, Kurt Keutzer, Dan Klein, and
  Joey Gonzalez.
\newblock Train big, then compress: Rethinking model size for efficient
  training and inference of transformers.
\newblock In {\em {ICML} 2020}.

\bibitem{DBLP:conf/acl/LiangZCJLHZC20}
Chen Liang, Simiao Zuo, Minshuo Chen, Haoming Jiang, Xiaodong Liu, Pengcheng
  He, Tuo Zhao, and Weizhu Chen.
\newblock Super tickets in pre-trained language models: From model compression
  to improving generalization.
\newblock In {\em {ACL/IJCNLP} 2021}.

\bibitem{DBLP:conf/nips/MichelLN19}
Paul Michel, Omer Levy, and Graham Neubig.
\newblock Are sixteen heads really better than one?
\newblock In {\em NeurIPS 2019}.

\bibitem{DBLP:conf/iclr/MolchanovTKAK17}
Pavlo Molchanov, Stephen Tyree, Tero Karras, Timo Aila, and Jan Kautz.
\newblock Pruning convolutional neural networks for resource efficient
  inference.
\newblock In {\em {ICLR} 2017}.

\bibitem{DBLP:conf/emnlp/PrasannaRR20}
Sai Prasanna, Anna Rogers, and Anna Rumshisky.
\newblock When {BERT} plays the lottery, all tickets are winning.
\newblock In {\em {EMNLP} 2020}.

\bibitem{radford2019language}
Alec Radford, Jeff Wu, Rewon Child, David Luan, Dario Amodei, and Ilya
  Sutskever.
\newblock Language models are unsupervised multitask learners.
\newblock 2019.

\bibitem{DBLP:conf/emnlp/RajpurkarZLL16}
Pranav Rajpurkar, Jian Zhang, Konstantin Lopyrev, and Percy Liang.
\newblock Squad: 100, 000+ questions for machine comprehension of text.
\newblock In {\em {EMNLP} 2016}.

\bibitem{sajjad2020poor}
Hassan Sajjad, Fahim Dalvi, Nadir Durrani, and Preslav Nakov.
\newblock Poor man's bert: Smaller and faster transformer models, 2020.

\bibitem{DBLP:journals/corr/abs-1910-01108}
Victor Sanh, Lysandre Debut, Julien Chaumond, and Thomas Wolf.
\newblock Distilbert, a distilled version of {BERT:} smaller, faster, cheaper
  and lighter.
\newblock {\em CoRR}, 2019.

\bibitem{DBLP:conf/aaai/ShenDYMYGMK20}
Sheng Shen, Zhen Dong, Jiayu Ye, Linjian Ma, Zhewei Yao, Amir Gholami,
  Michael~W. Mahoney, and Kurt Keutzer.
\newblock {Q-BERT:} hessian based ultra low precision quantization of {BERT}.
\newblock In {\em {AAAI} 2020}.

\bibitem{DBLP:conf/acl/SunYSLYZ20}
Zhiqing Sun, Hongkun Yu, Xiaodan Song, Renjie Liu, Yiming Yang, and Denny Zhou.
\newblock Mobilebert: a compact task-agnostic {BERT} for resource-limited
  devices.
\newblock In {\em {ACL} 2020}.

\bibitem{DBLP:conf/iclr/WangSMHLB19}
Alex Wang, Amanpreet Singh, Julian Michael, Felix Hill, Omer Levy, and
  Samuel~R. Bowman.
\newblock {GLUE:} {A} multi-task benchmark and analysis platform for natural
  language understanding.
\newblock In {\em {ICLR} 2019}.

\bibitem{DBLP:journals/corr/abs-1910-03771}
Thomas Wolf, Lysandre Debut, Victor Sanh, Julien Chaumond, Clement Delangue,
  Anthony Moi, Pierric Cistac, Tim Rault, R{\'{e}}mi Louf, Morgan Funtowicz,
  and Jamie Brew.
\newblock Huggingface's transformers: State-of-the-art natural language
  processing.
\newblock {\em CoRR}, 2019.

\bibitem{DBLP:conf/iclr/WuLLLH20}
Zhanghao Wu, Zhijian Liu, Ji~Lin, Yujun Lin, and Song Han.
\newblock Lite transformer with long-short range attention.
\newblock In {\em {ICLR} 2020}.

\bibitem{DBLP:journals/corr/abs-2002-02925}
Canwen Xu, Wangchunshu Zhou, Tao Ge, Furu Wei, and Ming Zhou.
\newblock Bert-of-theseus: Compressing {BERT} by progressive module replacing.
\newblock {\em CoRR}, 2020.

\bibitem{DBLP:conf/icml/YeGNZKL20}
Mao Ye, Chengyue Gong, Lizhen Nie, Denny Zhou, Adam Klivans, and Qiang Liu.
\newblock Good subnetworks provably exist: Pruning via greedy forward
  selection.
\newblock In {\em {ICML} 2020}.

\bibitem{DBLP:journals/corr/abs-1910-06188}
Ofir Zafrir, Guy Boudoukh, Peter Izsak, and Moshe Wasserblat.
\newblock {Q8BERT:} quantized 8bit {BERT}.
\newblock {\em CoRR}, 2019.

\end{thebibliography}

\appendix

  \section{Additional experiments and studies}
  \label{FirstAppendix}
  \subsection{AxFormer without using prior knowledge about tasks}
The use of prior knowledge about tasks is only a heuristic to help reduce the overheads of AxFormer. All the exact results in this paper can be obtained without the use of this prior knowledge. Prior research \cite{DBLP:conf/acl/JawaharSS19} has shown that the linguistic knowledge contained in different transformer layers can be demarcated into three main functional regions: phrase-level information in bottom layers (the layers closest to the input), semantic and syntactic information in the middle layers, and long-range dependency information in the top layers. Following these insights into the inner workings of transformer models, we consider all possible (3! = 6) orderings of these functional regions for pruning (Table \ref{tab:table41}), \emph{i.e.,} we create 6 different AxFormer models, each with elements inspected in different orders. Then, the AxFormer model with minimum validation loss is chosen as the final model for testing. The quality of the AxFormer model is worst when the blocks containing the most important linguistic knowledge are inspected first (bottom layers for sentiment analysis, see Appendix B), since these blocks may get pruned simply to alleviate the over-parameterization. In addition, we find that the ordering of elements within the same functional block (for example, elements in layer 10 before layer 11 or vice-versa) has negligible impact on the quality of the AxFormer model ($<$0.2 absolute points in all our studied tasks), further demonstrating the existence of regions with distinct types of linguistic knowledge. Without the use of prior knowledge, the average wall-clock time of AxFormer for BERT-Base increases from 7.6 minutes to 17.2 minutes on a single RTX 2080Ti GPU. The increase is not six-fold because in most orderings, the system enters ``bad" local minima very quickly. As a result, most of the elements inspected are deemed to be of high importance, and are not inspected at finer granularity (in fact, most orderings require less than 30 iterations in Alg. \ref{alg:alg5} for accuracy-driven pruning of BERT-Base models).

\begin{table}[htbp]

\caption{\textbf{AxFormer accuracy from inspecting elements in different orders on SST-2 using BERT-Base.} The baseline validation and dev accuracy of the fine-tuned model are 93.98 and 95.94, respectively.}

\begin{tabularx}{\linewidth}{@{} *3{>{\centering\arraybackslash}X}@{}}

\hline
\textbf{Order of inspection} & \textbf{Validation accuracy of AxFormer model} & \textbf{Test accuracy of AxFormer model}\\
\hline
Bottom, Middle, Top & 94.06 & 95.98\\

Bottom, Top, Middle & 94.38  & 96.01\\

Middle, Bottom, Top & 94.44 & 96.12\\

Middle, Top, Bottom & 95.22  & 96.28\\

Top, Bottom, Middle & 95.18 & 96.46\\

\textbf{Top, Middle, Bottom} & \textbf{95.24} & \textbf{96.7}\\
\hline
\end{tabularx}
\label{tab:table41}
\end{table}

\subsection{Analyzing the linguistic knowledge required for different downstream tasks}
Different downstream tasks require different types of linguistic knowledge. We find that the most important functional block of the transformer model for each task (corresponding to the most important kind of linguistic knowledge required to effectively solve the task) agrees with our intuition about the task. For example, we expect sentiment analysis to require only local context, since long-range information often ends up confusing the model (sentiments often change rapidly); it is also unlikely that syntactic and semantic information are needed. This is experimentally validated by the fact that inspecting the blocks in top layers, followed by the middle and finally the bottom layers, leads to most accurate models on SST-2 (Table \ref{tab:table40}). Similarly, we expect knowledge about language syntax and semantics to be most important for a task that tests the linguistic acceptability of a given sentence (CoLA). We also find that this holds across all transformer architectures that we study in this work (BERT, DistilBERT, Q8BERT and XLNet). We leverage this intuition to reduce the overheads of AxFormer without compromising on the quality of the generated models.

\begin{table}[htbp]

\caption{\textbf{The order of inspected elements that provides maximum accuracy for different downstream tasks using BERT, DistilBERT, Q8BERT and XLNet.} We find that the same ordering provides best performance on all the studied transformer architectures.}

\begin{tabularx}{\linewidth}{@{} *4{>{\centering\arraybackslash}X}@{}}

\hline
\textbf{Dataset} & \textbf{Task} & \textbf{Best ordering}& \textbf{Most important knowledge}\\
\hline
CoLA & Linguistic acceptability & Top, Bottom, Middle & Semantic/ Syntactic \\
\hline
MNLI & Entailment & Top, Middle, Bottom & Phrase-level \\
\hline
MRPC & Semantic equivalence & Top, Middle, Bottom & Phrase-level \\
\hline
QNLI & Question answering & Top, Middle, Bottom & Phrase-level \\
\hline
QQP & Semantic equivalence & Top, Middle, Bottom & Phrase-level \\
\hline
RTE & Entailment & Top, Middle, Bottom & Phrase-level \\
\hline
SST-2 & Sentiment Analysis & Top, Middle, Bottom & Phrase-level \\
\hline
STS-B & Sentence similarity & Top, Middle, Bottom & Phrase-level \\
\hline
WNLI & Reading comprehension & Top, Middle, Bottom & Phrase-level \\
\hline
SQUAD & Question answering & Top, Middle, Bottom & Phrase-level \\
\hline
\end{tabularx}
\label{tab:table40}
\end{table}

\subsection{Analyzing the benefits of AxFormer}
Transformers are pre-trained on a large text corpus through a self-supervised NLP task, such as predicting the next word in text given all the preceding words. When the pre-trained models are fine-tuned for different downstream tasks, they contain  task-irrelevant information that adds noise and confuses the model. In addition, since the pre-trained models are highly overparameterized, they severely overfit the small fine-tuning datasets. AxFormer helps the model focus on the task at hand by eliminating the irrelevant information in the fine-tuned models. 

\subsubsection{Improved Generalization}
\begin{figure}[htbp]
\centerline{\includegraphics[width=\linewidth]{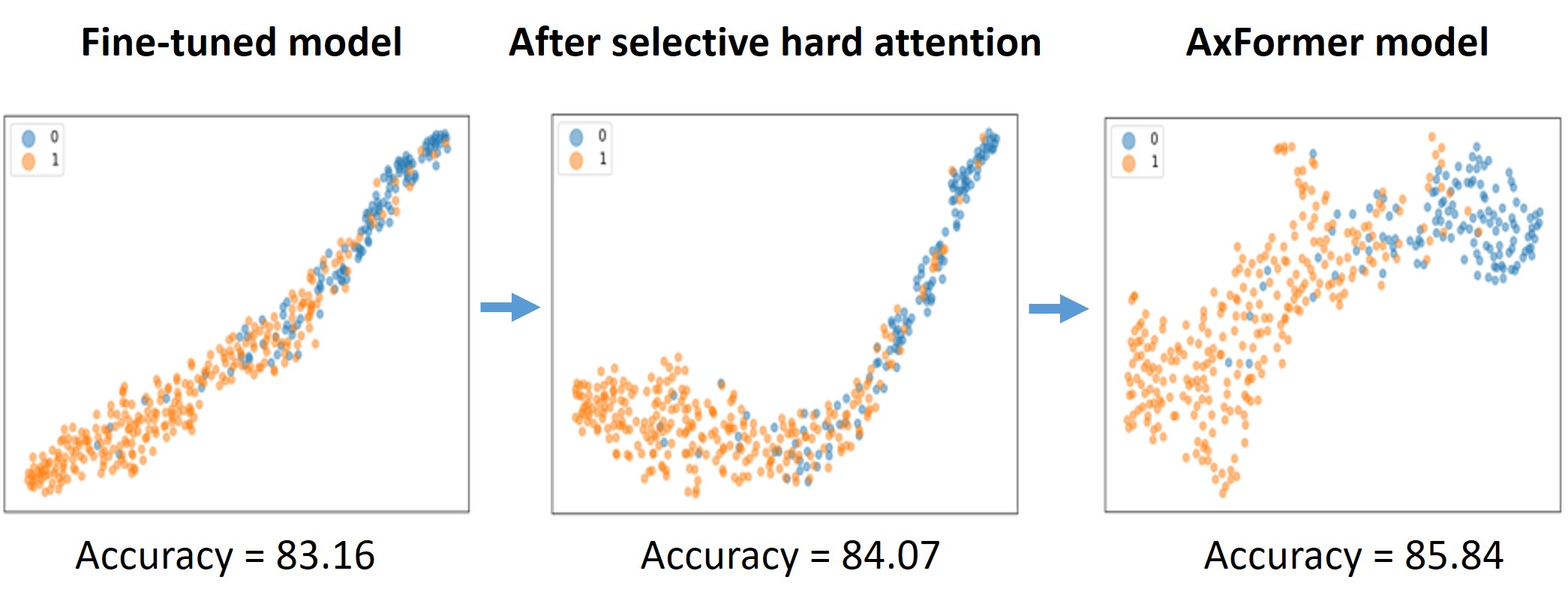}}
\caption{\textbf{Change in class boundaries using AxFormer on MRPC with BERT-Base.} We use T-distributed Stochastic Neighbor Embedding (TSNE) on word embeddings before the final classifier, using the 3 main components of each embedding. Here, blue and orange points refer to sentence pairs that are semantically non-equivalent and equivalent, respectively. AxFormer models show better class separability, and hence, are more accurate.}
\label{fig29}
\end{figure}

Accuracy-driven pruning can be seen as a form of training, since the objective of accuracy-driven pruning -- minimizing loss on the (validation) dataset -- is exactly the same as the objective of training. When large, over-parameterized models are trained on very limited data, the constraints of accuracy-driven pruning -- (1) loss can be reduced only be setting weights to 0 (equivalent to pruning elements), and no other changes to weights are allowed, and (2) a majority of samples in the validation set must benefit from each weight update; reduction in loss is a necessary but insufficient condition for weights to get updated -- help it learn more effectively than SGD. As a result, we find that training with SGD on a subset of the training data, followed by accuracy-driven pruning on the remaining unseen data, produces models with better generalization performance than training with SGD on the entire dataset (Figure \ref{fig29}). 

\subsubsection{Reduced variance}
Accuracy-driven pruning and selective hard attention filter out noise in the model by removing irrelevant information that ends up confusing the model. Accuracy-driven pruning reduces noise by pruning task-irrelevant elements. Selective hard attention further reduces noise by ensuring that the model focuses only on the relevant parts of the input. We find that this, in turn, leads to reduced cross-seed variance on the test set. However, the reduction in variance is limited by the effects of random initialization of the task-specific final layer.


\end{document}